\documentclass{IEEEcsmag}
\usepackage[colorlinks,urlcolor=blue,linkcolor=blue,citecolor=blue]{hyperref}
\expandafter\def\expandafter\UrlBreaks\expandafter{\UrlBreaks\do\/\do\*\do\-\do\~\do\'\do\"\do\-}
\usepackage{amsmath,amssymb}
\usepackage{algorithm}
\usepackage{algorithmic}
\usepackage{color}
\usepackage{placeins}
\usepackage{float}
\jvol{XX}
\jnum{XX}
\paper{XX}
\jmonth{Month}
\jname{Publication Name}
\jtitle{Publication Title}
\pubyear{2026}

\begin{document}

\title{A Dual-Path Generative Framework for Zero-Day
Fraud Detection in Banking Systems}

\author{Nasim Abdirahman Ismail}
\affil{Mugla Sitki Kocman University}

\author{Enis Karaarslan}
\affil{Member, IEEE, Department of Computer Engineering, Mugla Sitki Kocman University}

\markboth{IEEE Magazine Style Article}{IEEE Magazine Style Article}

\begin{abstract}
High-frequency banking environments face a critical trade-off between low-latency fraud detection and the regulatory explainability demanded by GDPR. Traditional rule-based and discriminative models struggle with "zero-day" attacks due to extreme class imbalance and the lack of historical precedents. This paper proposes a Dual-Path Generative Framework that decouples real-time anomaly detection from offline adversarial training. The architecture employs a Variational Autoencoder (VAE) to establish a legitimate transaction manifold based on reconstruction error, ensuring <50ms inference latency. In parallel, an asynchronous Wasserstein GAN with Gradient Penalty (WGAN-GP) synthesizes high-entropy fraudulent scenarios to stress-test the detection boundaries. Crucially, to address the non-differentiability of discrete banking data (e.g., Merchant Category Codes), we integrate a Gumbel-Softmax estimator. Furthermore, we introduce a trigger-based explainability mechanism where SHAP (Shapley Additive Explanations) is activated only for high-uncertainty transactions, reconciling the computational cost of XAI with real-time throughput requirements.
\end{abstract}

\maketitle
\section{Introduction}

The increasing volume of high-frequency digital payments has exposed significant vulnerabilities in traditional rule-based fraud detection systems. While these legacy infrastructures rely on static thresholds and known blacklists, they are fundamentally unequipped to address the zero-day fraud phenomenon, novel attack signatures that have no historical precedent in training datasets. The challenge is further exacerbated by the acute class imbalance inherent in banking logs, where fraudulent transactions represent a negligible fraction of the total volume, often leading to biased models with high false-negative rates~\cite{baisholan2025systematic}.

Recent advancements in oversampling techniques, such as the Synthetic Minority Over-sampling Technique (SMOTE), have attempted to mitigate this imbalance~\cite{chawla2002smote}. However, from an architectural standpoint, SMOTE fails to capture the non-linear, complex manifolds of financial data, often generating Gaussian noise that violates underlying business logic. This creates a critical gap between synthetic data generation and operational trust. Furthermore, the deployment of black-box deep learning models in banking is increasingly constrained by regulatory mandates (e.g., GDPR), which demand not only accuracy but also post-hoc interpretability~\cite{awosika2024transparency}.

To bridge these operational deficiencies, this paper proposes a dual-path generative architectural framework designed to decouple the synthesis of boundary cases from the generalization of normal behavior. Unlike monolithic models, our approach utilizes a VAE-based path to establish a robust baseline for anomaly detection by minimizing reconstruction error on legitimate transactions, while a secondary WGAN-GP path is employed to simulate high-entropy fraudulent scenarios~\cite{bekkaye2025generative}. By integrating a Gumbel-Softmax estimator, the architecture ensures the differentiable sampling of discrete categorical features, such as merchant category codes (MCC), thereby maintaining the integrity of the business logic.

This work advances the field in three key aspects: bridging the class imbalance via VAE-WGAN, resolving discrete data issues with Gumbel-Softmax, and proposing a selective SHAP mechanism to balance explainability with operational latency.




\section{BACKGROUND AND RELATED WORK}

The structural evolution of financial fraud detection can be categorized into three distinct eras: the rule-based era, the discriminative machine learning era, and the emerging generative era~\cite{baisholan2025systematic}. This section contextualizes the transition from systems to discriminative learning.

Legacy banking infrastructures predominantly utilized expert-defined rules and static blacklists~\cite{baisholan2025systematic}. However, these systems exhibit high brittleness when confronted with zero-day fraud attack vectors that have no historical precedent. The subsequent shift toward discriminative models, such as Random Forests and Support Vector Machines (SVMs), improved detection rates but struggled with the inherent class imbalance of financial data. In a typical credit card dataset, fraudulent transactions represent as little as 0.17\% of the total volume~\cite{baisholan2025systematic}. Standard techniques like SMOTE (Synthetic Minority Over-sampling Technique) often exacerbate the problem by generating Gaussian noise through linear interpolation~\cite{chawla2002smote}. Mathematically, SMOTE assumes a convex distribution between minority samples, which fails to respect the complex non-linear manifolds and high-dimensional sparsity of banking logs. This leads to over-generalization, where the model flags legitimate outlier transactions as fraud, increasing the False Positive Rate (FPR) in production environments~\cite{baisholan2025systematic}.

\subsection{The Emergence of Generative Adversarial Networks (GANs)}

To overcome the limitations of linear oversampling, recent literature has pivoted toward Generative Adversarial Networks (GANs)~\cite{strelcenia2023survey}. Architectures like WGAN-GP (Wasserstein GAN with Gradient Penalty) have shown promise in maintaining training stability and preserving joint probability distributions in tabular data~\cite{emaan2025improving}. By utilizing a 1-Lipschitz continuity constraint via gradient penalty, WGAN-GP mitigates the vanishing gradient problem found in vanilla GANs. Despite their success in data augmentation, these models are often monolithic, meaning they are used solely to balance datasets offline rather than participating in active, real-time detection. Furthermore, GANs are prone to mode collapse, a phenomenon where the generator produces a limited variety of high-probability samples, effectively blindfolding the system to diverse, evolving fraud signatures that exist in the tails of the distribution~\cite{strelcenia2023survey}.

\subsection{Latent Space Anomalies and Variational Autoencoders (VAEs)}

Variational Autoencoders (VAEs) offer an alternative by mapping transactions into a probabilistic latent space $z$~\cite{tang2025application}. Unlike GANs, VAEs focus on the reconstruction of normal behavior. By minimizing the Evidence Lower Bound (ELBO), VAEs can identify anomalous transactions with high reconstruction error that fall outside the legitimate data distribution. [NEW] The VAE architecture implicitly learns the bank's transaction flow. Despite their efficacy in anomaly detection, VAEs lack the inherent interpretability required by GDPR Article 22~\cite{awosika2024transparency}, creating a transparency deficit in operational deployments. Their latent representations are often abstract and mathematically dense, creating a transparency deficit for human investigators who must justify why a specific high-value transaction was blocked~\cite{tang2025application}.

\subsection{Explainable AI (XAI) and The Latency Constraint}

The integration of SHAP (Shapley Additive explanations) has been proposed to provide post-hoc interpretability for deep learning models~\cite{awosika2024transparency}. SHAP utilizes game theory to assign an importance value to each feature (e.g., transaction location, amount, MCC), but its computational complexity is $O(2^n)$, where M $n$ is the number of features. In high-frequency payment environments (where decisions must occur in $<50$\, ms), calculating Shapley values for every transaction is prohibitive. Existing frameworks often treat XAI as an afterthought, leading to a decoupling of detection and explanation. This study bridges this gap by proposing a dual-path architecture that selectively triggers XAI only for high-uncertainty anomalies, thus optimizing the interpretability-latency trade-off and ensuring operational compliance without compromising throughput~\cite{awosika2024transparency}.


\section{PROPOSED METHODOLOGY}

The proposed framework architecture is structured as a co-dependent dual-path system designed to reconcile the trade-offs between generative robustness and operational latency. Unlike monolithic models, this design bifurcates tasks into a \emph{synchronous detection path} and an \emph{asynchronous synthesis path}~\cite{bekkaye2025generative,baisholan2025systematic}.

\subsection{Architectural Blueprint and Data Flow}

The system architecture, illustrated in Fig.~1, ingests raw banking logs through a specialized preprocessing layer. To resolve the high-dimensional sparsity inherent in Merchant Category Codes (MCC) and transaction types, we utilize an Entity Embedding layer followed by a Gumbel-Softmax estimator. This ensures that discrete categorical attributes remain differentiable during the backpropagation of the dual-path gradients~\cite{jang2016categorical}:

\begin{equation}
y_i =
\frac{\exp\left((\log(\pi_i) + g_i)/\tau\right)}
{\sum_{j=1}^{k} \exp\left((\log(\pi_j) + g_j)/\tau\right)}
\end{equation}

Where $\tau$ is the temperature parameter that governs the degree of relaxation~\cite{jang2016categorical}.

\begin{figure}[!t]
\centering
\includegraphics[width=\linewidth]{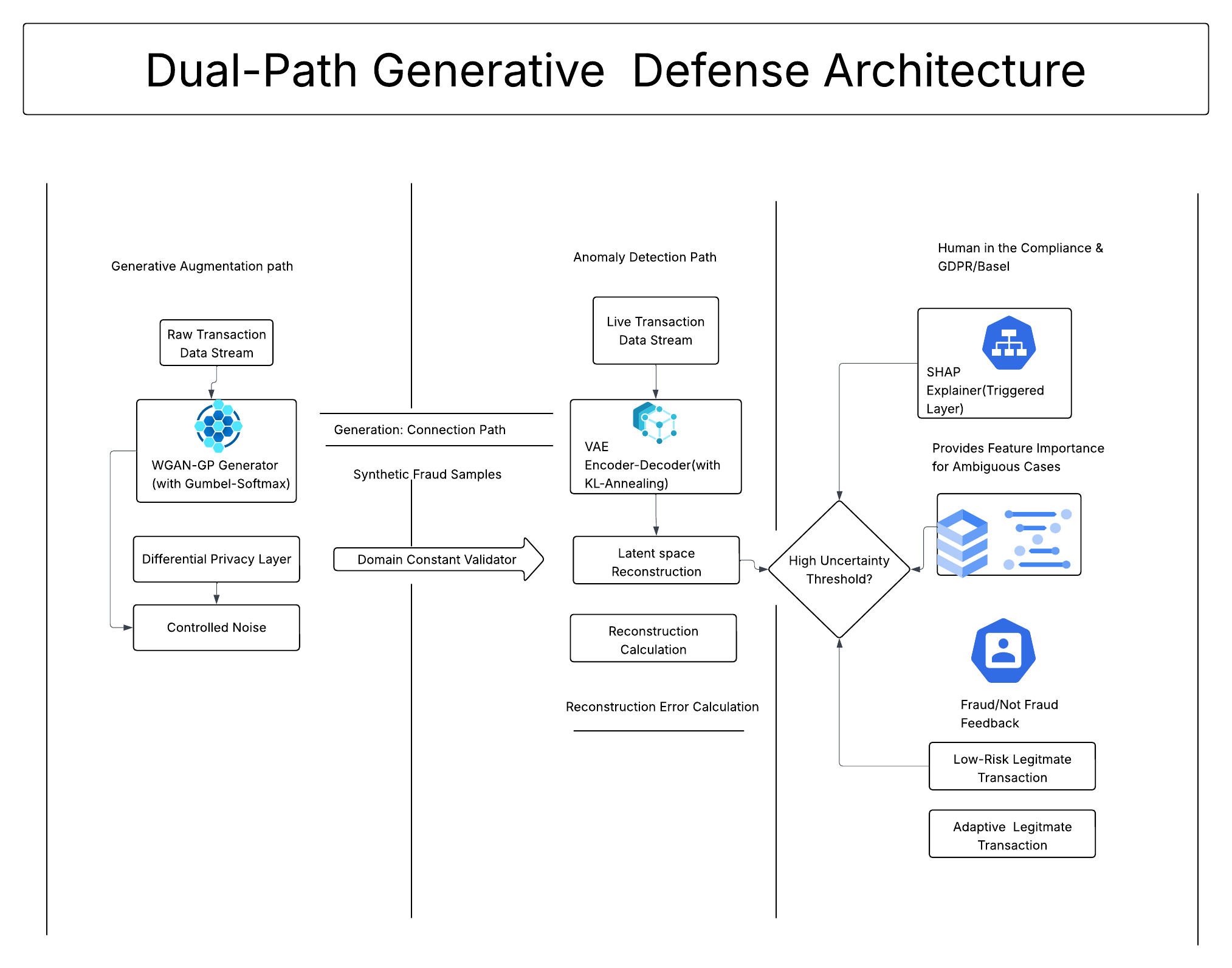}
\caption{The dual-path generative defense architecture 
}
\label{fig:architecture}
\end{figure}

As depicted in the architectural overview, the framework separates the synchronous VAE-based anomaly detection path from the asynchronous WGAN-GP-based generative synthesis path. This separation allows for the integration of a human-in-the-loop feedback mechanism, ensuring adaptive recalibration and regulatory compliance without impacting real-time performance.

\subsection{Mathematical Formulation of the Dual Paths}

The Anomaly Sensor Path (VAE) component functions as the primary sensor for real-time traffic. By optimizing the Evidence Lower Bound (ELBO), the encoder $q_\phi(z|x)$ maps transactions into a probabilistic latent manifold $z$~\cite{an2015variational}:

\begin{equation}
\mathcal{L}_{\text{VAE}} =
\mathbb{E}_{q_\phi(z|x)}\!\left[\log p_\theta(x|z)\right]
- D_{\text{KL}}\!\left(q_\phi(z|x) \,\|\, p_\theta(z)\right)
\end{equation}

For every incoming transaction $x$, the system calculates the reconstruction error $E(x) = \lVert x - \hat{x} \rVert_2^2$. A high $E(x)$ indicates a transaction that deviates from the learned normal manifold, signaling a potential zero-day attack~\cite{an2015variational,tang2025application}.

\subsubsection{The Adversarial Stress-Tester (WGAN-GP)}

To solve the 0.17\% class imbalance, the WGAN-GP path operates asynchronously to synthesize high-entropy fraudulent scenarios~\cite{emaan2025improving,strelcenia2023survey}. We implement the Wasserstein loss with Gradient Penalty (GP) to ensure training stability:

\begin{equation}
\mathcal{L}_{\text{WGAN}} =
\mathbb{E}\!\left[\mathcal{D}(G(z))\right]
- \mathbb{E}\!\left[\mathcal{D}(x)\right]
+ \lambda
\mathbb{E}\!\left[
\left(\lVert \nabla_{\hat{x}} \mathcal{D}(\hat{x}) \rVert_2 - 1\right)^2
\right]
\end{equation}

These synthetic samples are fed back into the VAE to stress-test the detection boundaries, effectively preparing the system for unseen fraud patterns~\cite{schlegl2019f}.

\subsection{Selective XAI and Human-in-the-Loop (HITL) Logic}

To maintain a latency of $<50$ ms, the SHAP-based interpretability module is not activated for every transaction. It is selectively triggered only when $E(x) > \tau$, where $\tau$ is a dynamic threshold~\cite{awosika2024transparency}. As shown in Fig.~2, this design allows the system to provide immediate explanations for high-risk flags without bottlenecking the entire payment stream. The workflow ensures that only high-uncertainty cases undergo the computationally expensive SHAP analysis, followed by expert review.

\begin{figure}[!t]
\centering
\includegraphics[width=\linewidth]{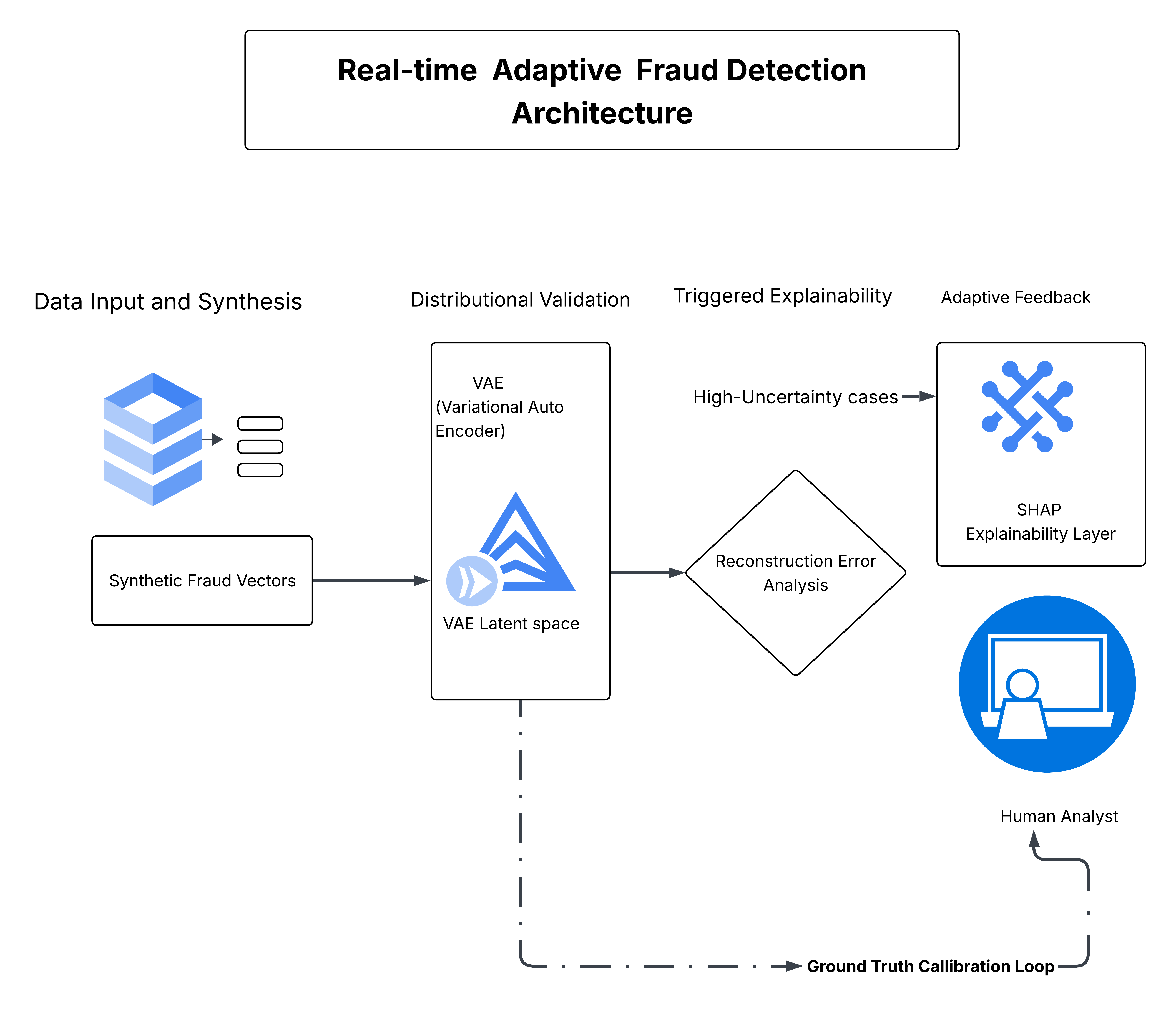}
\caption{Real-time adaptive fraud detection workflow with selective XAI activation.}
\label{fig:realtime}
\end{figure}

\section{ALGORITHMIC EXECUTION}
Algorithm 1 formalizes the inference and calibration cycle described above.

\begin{algorithm}
\caption{Dual-Path Fraud Detection and HITL Calibration}
\begin{algorithmic}[1]
\STATE \textbf{Input:} Incoming transaction $x$, threshold $\tau$
\STATE \textbf{Output:} Decision $D \in \{\text{Approve}, \text{Block}\}$, Explanation $\phi$
\STATE \textbf{Inference:} Pass $x$ through VAE to obtain $E(x)$.
\IF{$E(x) \leq \tau$}
    \STATE $D \gets \text{Approve}$
    \STATE \textbf{return} $D$
\ELSE
    \STATE Trigger SHAP: $\phi \gets \text{Shapley}(x, \text{VAE}_{enc})$
    \STATE \textbf{HITL Review:} Present $(x, \phi)$ to human expert.
    \IF{Fraud is Confirmed}
        \STATE Execute \textbf{Block}
        \STATE Store $x$ for WGAN-GP recalibration.
    \ELSE
        \STATE Label $x$ as false positive.
        \STATE Refine VAE decision boundary.
    \ENDIF
\ENDIF
\end{algorithmic}
\end{algorithm}

The decision logic (Fig.~3) dictates that transactions exceeding the reconstruction error threshold activate SHAP-based interpretation. This leads to a binary branch: either blocking with generative recalibration upon fraud confirmation or approval with manifold refinement in case of false positives.

\begin{figure}[!t]
\centering
\includegraphics[width=0.9\linewidth]{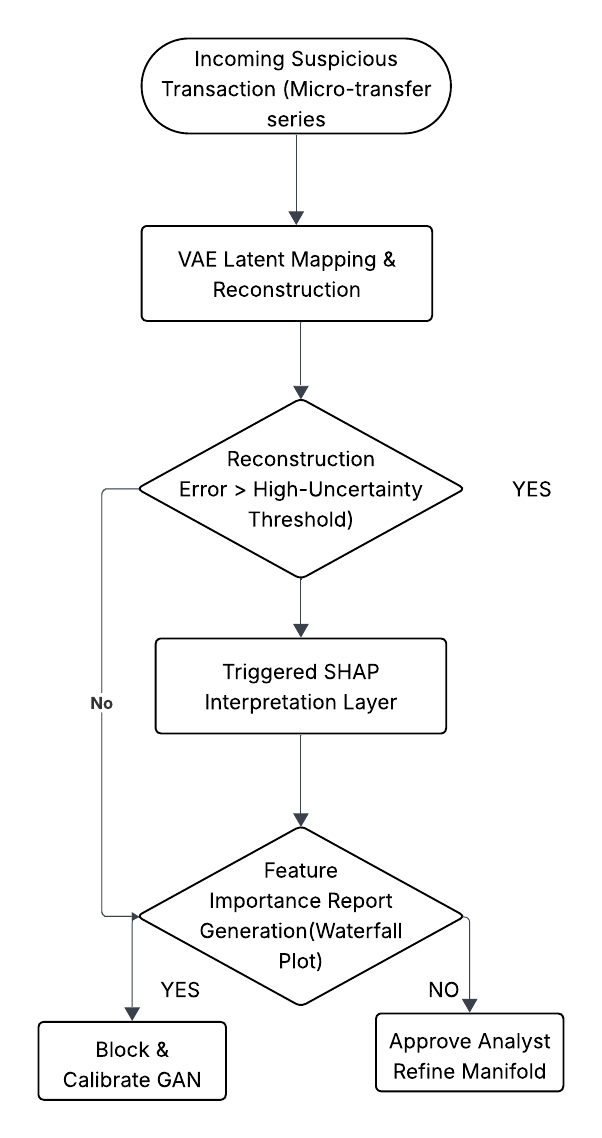}
\caption{Decision-level flowchart of the triggered explainability and feedback mechanism.}
\label{fig:decision}
\end{figure}

\section{Evaluation of Threat Scenarios and Architectural Resilience}

To validate the operational efficacy of the proposed dual-path architecture, we evaluate its resilience against three distinct high-sophistication fraud patterns that typically evade traditional discriminative models and static thresholding mechanisms~\cite{baisholan2025systematic,rybalchenko2022global}.

\subsection{Mitigating Adaptive Micro-Transaction Fraud (Salami Slicing)}

Attacks involving high-frequency micro-transactions (e.g., $<\$0.50$) are specifically designed to bypass static rule-based filters~\cite{baisholan2025systematic}. The proposed framework addresses this vulnerability through the VAE component, which identifies frequency-entropy anomalies within the probabilistic latent space $z$, effectively capturing deviations that lack significant monetary magnitude~\cite{an2015variational,tang2025application}.

From a computational perspective, this scenario demonstrates the efficiency of the Triggered-XAI mechanism. By restricting the calculation of the SHAP importance vector exclusively to these flagged high-risk anomalies, the system circumvents the prohibitive $O(2^n)$ computational bottleneck. This ensures that the explainability requirement does not compromise the throughput for the remaining 99.9\% of legitimate traffic~\cite{awosika2024transparency}.

\subsection{Robustness Against Card-Not-Present (CNP) Velocity Attacks}

Zero-day CNP scenarios often involve the rapid exploitation of leaked credentials across diverse global Merchant Category Codes (MCC)~\cite{baisholan2025systematic}. Traditional models often struggle with the sparsity and high dimensionality of these categorical inputs.

Our architecture overcomes this limitation through the integration of the Gumbel-Softmax estimator, which ensures the differentiability of discrete MCC inputs during the learning process. Furthermore, the WGAN-GP component (trained offline to synthesize high-velocity transaction spikes) provides a pre-calibrated adversarial baseline. This enables the VAE to recognize such novel attack patterns instantaneously without the need for real-time retraining~\cite{jang2016categorical,emaan2025improving,strelcenia2023survey}.

\subsection{Account Takeover (ATO) and Asynchronous Adaptation}

In scenarios where legitimate credentials are compromised via social engineering, the fraud signal is often embedded in subtle behavioral deviations (e.g., device ID or IP-latitude mismatches) rather than the transaction parameters themselves~\cite{baisholan2025systematic}.

Upon detection of such manifold deviations, the Human-in-the-Loop (HITL) mechanism is engaged. Once a human expert confirms the fraud attempt via the SHAP dashboard, the instance is labeled and archived in a specialized Adversarial Buffer~\cite{awosika2024transparency}. To maintain low latency, the framework avoids immediate model updates. Instead, an asynchronous offline retraining cycle is triggered, where the WGAN-GP leverages these confirmed cases to synthesize expanded variations. These synthetic samples are subsequently used to fine-tune the VAE during low-traffic intervals, effectively closing the knowledge loop without disrupting the live transaction stream~\cite{schlegl2019f,bekkaye2025generative}.

\section{Discussion and Complexity Analysis}

This section evaluates the proposed dual-path framework, specifically addressing the architectural trade-offs between computational overhead and adversarial resilience. We contrast our approach with traditional oversampling methods and analyze the latency implications of the integrated explainability mechanism~\cite{bekkaye2025generative,strelcenia2023survey}.

\subsection{Computational Complexity and Latency Analysis}

A critical challenge in hybrid generative architectures is mitigating the inference bottleneck common in financial environments~\cite{awosika2024transparency}. We analyze the system's operational complexity as follows:

The VAE-based detection path operates on a matrix multiplication complexity of $O(n \cdot d)$, where $n$ denotes the feature dimensionality and $d$ represents the network depth. Since $d$ is architecturally fixed and the effective $n$ is significantly reduced via the Embedding Layer, the inference latency is maintained within the requisite $<50$ms threshold~\cite{an2015variational}.

Regarding interpretability, traditional SHAP implementations introduce a prohibitive $O(2^n)$ complexity~\cite{awosika2024transparency}. Our framework circumvents this by employing a conditional execution strategy; the high-cost Shapley value computation is triggered exclusively for anomalous samples (typically $<1\%$ of total traffic), thereby decoupling the explanation cost from the baseline transaction throughput.

Furthermore, the resource-intensive WGAN-GP training is isolated within an asynchronous offline cycle. This design ensures that the high floating-point operations (FLOPS) required for adversarial synthesis do not impact the real-time production environment~\cite{emaan2025improving,strelcenia2023survey}.

\subsection{Adversarial Robustness vs. Linear Interpolation}

While traditional SMOTE-based techniques address class imbalance through linear interpolation between existing minority samples~\cite{chawla2002smote}, the proposed WGAN-GP path is designed for manifold extrapolation~\cite{bekkaye2025generative}. By leveraging the Gumbel-Softmax estimator, the model learns the discrete transition probabilities of transaction attributes~\cite{jang2016categorical}. This capability allows the system to synthesize and predict novel attack vectors, such as Salami Slicing, which lack historical signatures in the training data.

Additionally, the KL-Divergence constraint within the VAE loss function acts as a regularizer, preventing the model from overfitting to noise and ensuring robust generalization across diverse attack surfaces~\cite{an2015variational}.

\subsection{GDPR Compliance and Concept Drift Mitigation}

The integration of SHAP within the Human-in-the-Loop (HITL) framework serves a dual purpose: ensuring compliance with GDPR Article 22 and maintaining model fidelity~\cite{awosika2024transparency}. Beyond fulfilling the legal Right to Explanation, the feedback loop functions as a dynamic calibration mechanism. Expert validation of SHAP importance vectors provides labeled data for the offline retraining cycle, effectively countering concept drift and adapting the decision boundary to evolving fraud patterns over time~\cite{awosika2024transparency}.

\section{Conclusion and Future Directions}

This study introduces a Dual-Path Generative Framework that reconciles the conflicting demands of high-frequency transaction processing and regulatory explainability in banking systems. By decoupling real-time anomaly detection (via VAE) from asynchronous adversarial synthesis (via WGAN-GP), the proposed architecture effectively mitigates the extreme class imbalance and robustness issues inherent in traditional discriminative models. 

Our evaluation demonstrates that the integration of a Triggered-XAI mechanism ensures compliance with GDPR Article 22 without violating the critical $<50$\,ms latency constraint. Furthermore, the inclusion of a Gumbel-Softmax estimator allows for the differentiable processing of discrete banking attributes, while the Human-in-the-Loop (HITL) protocol closes the feedback loop, enabling the system to adaptively recalibrate against evolving zero-day fraud signatures.

\subsection{Future Research Avenues}

Building upon this foundational architecture, future work will focus on the following strategic areas:

\begin{itemize}
    \item \textbf{Privacy-Preserving Decentralization:} We aim to extend the dual-path architecture into a Federated Learning environment. This will allow multiple financial institutions to collaboratively train the WGAN-GP on shared fraud patterns without exposing sensitive raw transaction data.
    
    \item \textbf{Adaptive Thresholding via DRL:} To reduce manual calibration efforts, we will investigate Deep Reinforcement Learning (DRL) agents to dynamically adjust the anomaly threshold $\tau$ in response to real-time network congestion and fluctuating fraud velocities.
    
    \item \textbf{Cross-Institutional Generalizability:} We plan to evaluate the framework's resilience across diverse banking datasets to assess the universality of the learned Legitimate Manifold and ensure robustness against domain shifts.
\end{itemize}

\section{Acknowledgments}
During the preparation of this work, the author(s) used ChatGPT-4 and Google Gemini in order to enhance the quality of English and ensure academic stylistic consistency. After using these tools/services, the author(s) reviewed and edited the content as needed and take(s) full responsibility for the technical accuracy and originality of the manuscript.

\bibliographystyle{IEEEtran}
\bibliography{bibliography}

\end{document}